\documentclass[10pt,twocolumn,letterpaper]{article}

\usepackage{iccv}              % To produce the CAMERA-READY version
\definecolor{iccvblue}{rgb}{0.21,0.49,0.74}
\usepackage[pagebackref,breaklinks,colorlinks,allcolors=iccvblue]{hyperref}
\usepackage{multirow}
\usepackage{svg}
\usepackage{tabularx}
\usepackage{xcolor}  % 提供颜色功能
\usepackage{lipsum}  % 演示用，生成示例文本
\usepackage{graphicx}

\newcommand{\redbf}[1]{\textbf{\textcolor{red}{#1}}}
\newcommand{\bluebf}[1]{\textbf{\textcolor{blue}{#1}}}
\def\confName{ICCV}
\def\confYear{2025}

\title{DSPO: Direct Semantic Preference Optimization for Real-World Image Super-Resolution}
\author{Miaomiao Cai$^{1}$\footnotemark[1]\quad
Simiao Li$^{2}$\footnotemark[1]\quad Wei Li$^{2}$\footnotemark[2]\quad 
Xudong Huang$^{2}$\quad 
Hanting Chen$^{2}$\quad 
Jie Hu$^{2}$\quad 
Yunhe Wang$^{2}$\footnotemark[2]\\
$^1$University of Science and Technology of China\\
$^2$Huawei Noah’s Ark Lab\\
\tt\small mmcai@mail.ustc.edu.cn, lisimiao@huawei.com, wei.lee@huawei.com
}

\begin{document}
\maketitle
\renewcommand{\thefootnote}{\fnsymbol{footnote}}
\footnotetext[1]{Equal Contribution}
\footnotetext[2]{Corresponding Author}

\begin{abstract}
Recent advances in diffusion models have improved Real-World Image Super-Resolution (Real-ISR), but existing methods lack human feedback integration, risking misalignment with human preference and may leading to artifacts, hallucinations and harmful content generation.
To this end, we are the first to introduce human preference alignment into Real-ISR, a technique that has been successfully applied in Large Language Models and Text-to-Image tasks to effectively enhance the alignment of generated outputs with human preferences.  
Specifically, we introduce Direct Preference Optimization (DPO) into Real-ISR to achieve alignment, where DPO serves as a general alignment technique that directly learns from the human preference dataset.
Nevertheless, unlike high-level tasks, the pixel-level reconstruction objectives of Real-ISR are difficult to reconcile with the image-level preferences of DPO, which can lead to the DPO being overly sensitive to local anomalies, leading to reduced generation quality.
To resolve this dichotomy, we propose Direct Semantic Preference Optimization (DSPO) to align instance-level human preferences by incorporating semantic guidance, which is through two strategies: (a) semantic instance alignment strategy, implementing instance-level alignment to ensure fine-grained perceptual consistency, and (b) user description feedback strategy, mitigating hallucinations through semantic textual feedback on instance-level images.
As a plug-and-play solution, DSPO proves highly effective in both one-step and multi-step SR frameworks.

\end{abstract}   
%liwei
\section{Introduction}
% 现实世界图像超分辨率（Real-ISR）的目标是从具有复杂和未知退化的低分辨率观测中重建人类感知上逼真的高分辨率图像。最近，得意于diffusion model在生成过程的稳定性和良好的图像生成质量，其在超分任务上取得了excellent的进展。The recently developed generative diffusion models (DM) [39, 16], especially the large-scale pre-trained text-to-image (T2I) models [37, 36], have demonstrated remarkable performance in various downstream tasks. Having been trained on billions of image-text pairs, the pre-trained T2I models possess powerful natural image priors, which can be well exploited to improve the naturalness and perceptual quality of Real-ISR outputs
% 当前diffusion-based SR model 通常使用pair对的LR-HR图像在a single stage进行end-to-end的supervised训练，which训练过程，没有人类反馈的参与。
%人工不干预的情况下，现有的模型训练范式优化目标和人类感知之间仍具有gap，可能导致潜在的有害内容，幻觉，伪纹理的产生。
Real-world Image Super-Resolution (Real-ISR)~\cite{chen2022real,zhang2023real,wang2024exploiting} aims to reconstruct photo-realistic high-quality (HQ) images from low-quality (LQ) images with various degradations such as noise, blur, and low-resolution. Recently, diffusion models~\cite{ho2020denoising,dhariwal2021diffusion,song2020score} have made excellent progress in Real-ISR tasks~\cite{wu2025one,wang2024exploiting,yang2024pixel,lin2024diffbir,wu2024seesr,yu2024scaling}, owing to their remarkable capability of generation. 
%显著的生成能力
%However, diffusion-based SR (SR) models typically use paired LQ-HQ images for end-to-end supervised training in a single stage, with no human feedback involved during the training process. 
%人工不干预的情况下，现有的模型训练范式优化目标和人类感知之间仍具有gap，可能导致潜在的有害内容，幻觉，伪纹理的产生。
%但是由于引入了diffusion loss, 任务的输出可能会产生不符合逻辑的伪纹理和乱生成现象.which 不可以通过监督的方法去解决在固定的数据集上. 为了解决这个问题,我们引入alignem技术
However, these models generally employ the supervised training paradigm that directly learns from paired LQ-HQ image datasets, omitting human feedback throughout the training cycle. Without human intervention, the optimization objectives of these models may misalign with human perceptual preferences, leading to potentially harmful content generation, hallucination phenomena, and visual artifacts.

 \begin{figure}[t]
  \centering
   \includegraphics[width=0.7\linewidth]{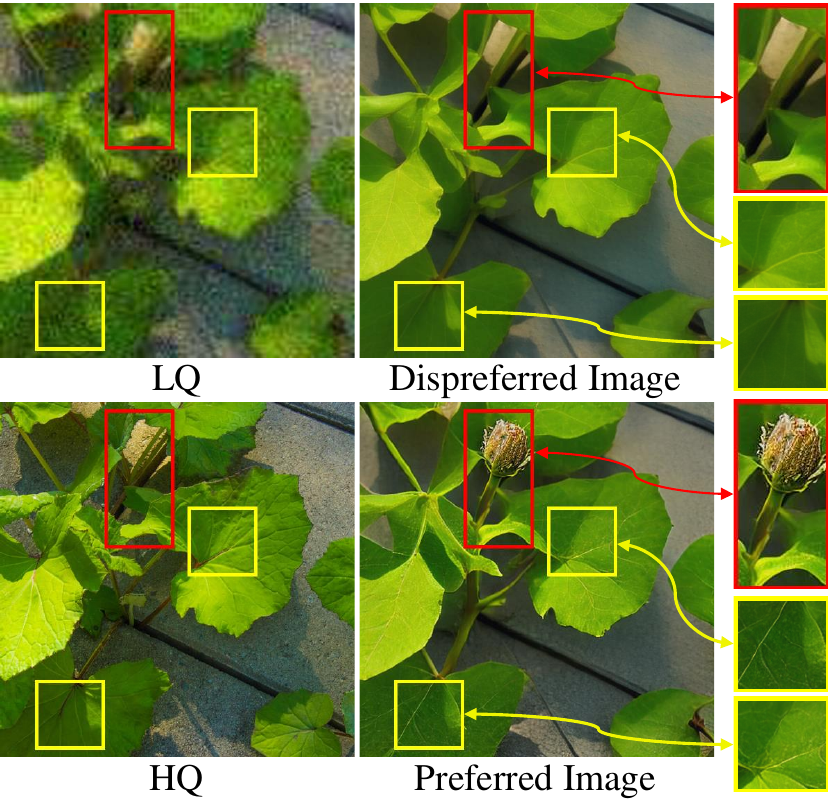}
   % \vspace{-3mm}
   \caption{The dilemma between image-level preferences in DPO and pixel-level reconstruction objectives in Real-ISR: The preferred image, selected by the `winner of overall image visual pleasure' rule, appears sharper in the yellow bbox and other areas but shows local hallucinations in the red bbox, where the dispreferred image performs better.
   % The dilemma between image-level preferences of existing DPO and the pixel-level reconstruction objectives of Real-ISR tasks. The preferred image, chosen based on the `winner of overall image visual pleasure' rule, appears sharper within the yellow bbox compared to the dispreferred one. However, it exhibits local hallucination that contradicts factual details in the red bbox, whereas the dispreferred image performs better in the red bbox. 
   %The drawback of image-level DPO in Real-ISR: When comparing the overall image, the preferred image generally exhibits high-quality effects, such as within the yellow box. However, in the red region, the preferred image tends to produce artifacts or uncontrolled generation.
   }
   % \vspace{-5mm}
   \label{fig:teaser}
\end{figure}
% 在LLM或T2I生成任务中，通过引入人类反馈对其，解决了xxx的问题。具体来说，是分两阶段去解决的。第一个（“预训练”）阶段在大规模的网络数据上进行训练。第二个（“对齐”）阶段通过微调使其更好地与人类偏好对齐。对齐通常通过监督微调（SFT）或者从人类反馈的强化学习（RLHF）来进行，使用偏好数据进行训练。
% Such misalignment between model outputs and human preferences also widely exists in other generative tasks.  For instance, in the fields of Large Language Models (LLMs)\cite{achiam2023gpt,touvron2023llama} and Text-to-Image (T2I) generation\cite{wallace2024diffusion,li2025aligning}, human preference alignment techniques have been widely employed to mitigate misalignment issues by fine-tuning the pre-trained model with Reinforcement Learning from Human Feedback (RLHF) strategies. RLHF can be broadly categorized into reward model-based and reward model-free methods. The former utilizes a reward model explicitly trained to predict human preferences, guiding policy optimization towards outputs aligned with human judgment, such as PPO~\cite{schulman2017proximal} and GRPO~\cite{shao2024deepseekmath} in LLMs, and DDPO~\cite{ho2020denoising} in T2I. This approach is particularly suitable for tasks with clearly defined reward signals, such as code generation and mathematical problem-solving. In contrast, the reward model-free method optimizes the policy model directly based on human preference data without training an explicit reward model, exemplified by methods like DPO~\cite{rafailov2023direct} in LLM and Diffusion-DPO~\cite{wallace2024diffusion} in T2I.
Such misalignment between model outputs and human preferences also exists in other tasks.  For instance, in the fields of Large Language Models (LLMs)\cite{achiam2023gpt,touvron2023llama} and Text-to-Image (T2I) generation\cite{wallace2024diffusion,li2025aligning}, human preference alignment techniques have been widely employed to mitigate misalignment issues by fine-tuning the pre-trained model with Reinforcement Learning from Human Feedback (RLHF) strategies. 
The classical RLHF paradigm (\textit{e.g.}, PPO~\cite{schulman2017proximal}, DDPO~\cite{ho2020denoising}) first trains a reward model on a fixed preference dataset, then optimizes the policy to maximize the predicted reward. However, relying on a reward model makes the process inherently complex and significantly increases computational overhead~\cite{rafailov2023direct}. 
In contrast, Direct Preference Optimization (DPO)~\cite{rafailov2023direct} is proposed to directly optimize the policy model based on human preference data without reward model and demonstrate excellent performance on generative tasks~\cite{rafailov2023direct,wallace2024diffusion}.  
Despite success in LLM and T2I, human preference alignment (\textit{e.g.} DPO) remains unexplored in Real-ISR. 
%We notice that training a reward model is challenging due to the complexity of human perception, which should consider factors like structural fidelity, contrast, realism, and style, with individual variability. 
% If we adhere to the classical RLHF training paradigm, the process first requires training a reward model based on a fixed preference dataset, followed by optimizing the policy to maximize the reward as predicted by this model. However, this entire pipeline is inherently complex and introduces additional computational overhead~\cite{rafailov2023direct}. 
Therefore, we introduce human preference alignment into Real-ISR for the first time through DPO.
%
% DPO is a general method for human preference alignment that trains on preferred and dispreferred data from humans, without training an extra reward model.
% which is particularly well-suited for SR tasks, \textcolor{red}{as training a reward model in SR is often challenging}. 
% However, our experiments indicate that directly applying DPO to diffusion-based SR frameworks results in performance degradation.
However, directly applying DPO to representative Real-ISR models results in performance degradation due to the inherent dilemma between image-level preferences of existing DPO and the pixel-level reconstruction objectives of Real-ISR tasks.
Specifically, as illustrated in Fig.~\ref{fig:teaser}, image-level preferences may lead to artifacts or hallucinations in preferred images within local regions, especially in high-frequency texture and complex regions. Such conflicts may lead to the models being overly sensitive to local anomalies, exhibiting fluctuations and ambiguity during training, which ultimately affects the quality of the generated performance.

% 为了解决这一限制，我们提出了Direct Semantic Preference Optimization (DSPO)， that deeply aligns instance-level human preferences by incorporating semantic guidance。具体来说，我们提出Semantic Instance Alignment Module来防止局部乱生成和伪影：首先利用instance segmentation model提取instance，然后依据人类的偏好选择instance-level的最优实例和最差实例，记录偏好，进行instance-level的DPO优化。
% 此外，为了进一步抑制乱生成现象，我们提出User descreption Feedback Module：我们利用VLM提取instance-level的语义信息，并根据人类反馈选择生成错误语义信息的实例作为负向提示，以抑制乱生成的现象。

To address this problem, we propose Direct Semantic Preference Optimization (DSPO), which deeply aligns instance-level human preferences by incorporating semantic guidance.
Specifically, we propose the semantic instance alignment strategy that conducts human preference alignment at the instance level to achieve finer-grained alignment. An instance semantic extraction model is employed to extract individual instances, and then preferred and dispreferred instance-level cases are selected and recorded, followed by instance-level preference alignment. 
% The experiment proves that this strategy significantly improves the alignment ability and generation quality of  Real-ISR models.
% address artifacts and hallucination in local regions: first, an instance semantic extraction model is employed to extract individual instances, and then preferred and dispreferred instance-level cases are selected and recorded, followed by instance-level DPO. 
%
Additionally, to further mitigate the hallucination phenomenon, we propose the user description feedback strategy: we incorporate users' semantic textual feedback on instance-level images and select hallucination semantic information texts as prompt injection.  
% The experiments demonstrate that this strategy avoids the hallucination phenomenon and further improves human preference alignment capability and generation ability.
%进一步提升了模型性能
% Additionally, to further mitigate the phenomenon of hallucination, we propose the User Description Feedback Module: we leverage Vision-Language Models (VLM) to extract instance-level semantic information and, based on human feedback, select hallucination semantic information at the instance level as negative prompts.

Our contributions are as follows:
\begin{enumerate}
\item %We introduce human preferences into the SR task, which is the first to integrate Reinforcement Learning with Human Feedback (RLHF) with SR.
We pioneer introducing human preference alignment into Real-ISR, establishing the first methodological approach to incorporate human preference alignment in this field.
\item We propose DSPO, which achieves instance-level human preference alignment and significantly suppresses artifacts and hallucination phenomena.
\item  As a plug-and-play solution, DSPO demonstrates significant effectiveness in both one-step and multi-step SR frameworks, achieving notable improvements in perception and fidelity metrics.
\end{enumerate}
% \vspace{-1mm}

\section{Related Work}
% \vspace{-1mm}
\subsection{Generative SR Models}
%需要强调一下为什么diffuion-based比gan-based要好，不然要加gan-based的实验
%%%%%%%%%%%%%%%%%%%%%%%%%%%%%%%%%%%%%%%%%%%%%%%%%%
% 超分辨率（SR, SR）任务旨在从低分辨率（LR）图像重建高分辨率（HR）图像，在医学影像、遥感、视频增强等领域具有广泛应用。传统 SR 方法主要依赖卷积神经网络（CNN）【1】和生成对抗网络（GAN）【2】优化像素重建与感知质量，而扩散模型（Diffusion Models, DM）凭借其强大的生成能力，在 SR 任务中展现出卓越性能【3】。
Traditional super-resolution (SR) methods rely on Convolutional Neural Networks (CNNs)~\cite{ledig2017photo} and Generative Adversarial Networks (GANs)~\cite{dong2014learning}  for pixel reconstruction and perceptual quality, while diffusion models demonstrate exceptional performance in SR tasks due to their robust generative capabilities~\cite{saharia2022image}.
Diffusion models restore high-quality images via stepwise denoising, becoming central to Real-ISR. Early diffusion ISR methods~\cite{kawar2022denoising} relied on Denoising Diffusion Probabilistic Models (DDPM) but struggled with complex degradations~\cite{song2020score}. Then, many approaches are proposed to tackle these challenges~\cite{yu2024scaling,yang2024pixel,wang2024exploiting,wu2024seesr}. SUPIR~\cite{yu2024scaling} leverages SDXL generative capabilities and LLaVA text understanding to enhance detail recovery. StableSR~\cite{wang2024exploiting} integrates a temporal-aware encoder to improve recovery quality, particularly in video scenarios. SeeSR~\cite{wu2024seesr} utilizes text-guided diffusion to enhance semantic consistency and detail in the generated images. Recent on-step diffusion SR models introduce efficient inference strategies to optimize diffusion for one-step SR. OSEDiff~\cite{wu2025one} employs Variational Score Distillation (VSD) to improve performance and reduce computational costs.
\subsection{Human Preference Alignment in LLMs}
To align human preferences, LLMs typically utilize supervised fine-tuning (SFT) following pre-training and then use RLHF to enhance model human alignment and generation quality~\cite{ouyang2022training}. 
The traditional RLHF guide policy optimization by training a reward model to generate outputs that align with human expectations, such as Proximal Policy Optimization (PPO)~\cite{schulman2017proximal}. However it faces challenges to train a reward model when reward signal is unclear, suffering from high computational costs and training instability.
To overcome the limitations, Direct Preference Optimization (DPO) serves as an alternative method, allowing LLMs to optimize directly based on pairwise preference data without training a reward model~\cite{rafailov2023direct}. DPO has low computational overhead and stable optimization, demonstrating superior performance on open-source models like Llama 2~\cite{bai2022training}. 
Compared to reward model-based methods, DPO is more efficient in optimizing LLM preferences, reducing training complexity while maintaining competitive performance~\cite{touvron2019open}.
%%%%%%%%%%%%%%%%%%%%%%%%%%%%%%%%%%%%%%%%%%%
% Large Language Models (LLMs) undergo extensive pre-training and are typically fine-tuned using Supervised Fine-Tuning (SFT) for initial optimization. Subsequently, they are aligned through Reinforcement Learning from Human Feedback (RLHF) to enhance controllability and alignment with user preferences. RLHF relies on a reward model (RM) to predict user preferences, and the language model is adjusted through policy optimization to maximize RM scores.
% Despite the success of RLHF in improving LLM quality, it faces challenges such as high computational costs, training instability, and potential optimization imbalances due to reward hacking.
% Direct Preference Optimization (DPO) emerges as an efficient alternative to RLHF, eliminating the reinforcement learning component and relying solely on pairwise preference data to optimize the model directly. DPO works by maximizing the probability of superior samples while suppressing inferior ones, thereby aligning the model with human preferences without the need to train an RM or perform policy optimization. Additionally, DPO employs KL constraints to ensure that the optimized model does not deviate significantly from the original distribution, thus maintaining generation quality while enhancing alignment capabilities.
% Compared to RLHF, DPO offers lower computational overhead, greater optimization stability, and improved generalization ability, positioning it as a significant direction for preference alignment in the current LLM landscape.
\subsection{Human Preference Alignment in T2I}
%DDPO Diffusion-dpo
% Text-to-Image (T2I) diffusion models are widely applied in the AIGC domain, relying on static data distributions to optimize generation quality~\cite{nichol2021glide}. However, existing methods struggle to adapt to user preferences, limiting the semantic consistency, artistic style, and aesthetic quality of the generated results~\cite{dhariwal2021diffusion}. 
Human preference alignment has emerged as a key direction for enhancing the subjective quality of T2I tasks. ImageReward~\cite{xu2023imagereward} trains reward models using human rating data to optimize generative preferences. However, this method is susceptible to bias and has limited generalization capabilities~\cite{bai2022training}.
% Direct Preference Optimization (DPO) serves as an alternative that does not require a reward model. 
% DDPO~\cite{ho2020denoising} optimize diffusion models within a small vocabulary range but struggle to adapt to complex text prompts. %这里能不能加一下DDPO是reward model带来的劣势
DDPO~\cite{ho2020denoising} optimizes diffusion models within a small vocabulary range but struggles to adapt to complex text prompts, highlighting the limitations associated with reward model-based approaches. 
In contrast, Diffusion-DPO fine-tunes diffusion models directly based on human preference data without requiring an explicit reward model~\cite{wallace2024diffusion}, enhancing the generation quality for open vocabulary without increasing inference costs, thereby aligning T2I tasks more closely with human aesthetics and semantic consistency.
\section{Preliminaries}
%直接偏好优化（DPO）是一种通过用户偏好来指导生成模型的优化方法。其核心思想是通过评估生成结果与用户反馈之间的相似度，提升模型生成内容的质量。
% Direct Preference Optimization (DPO) is a method for optimizing generative models guided by user preferences. Its core idea is to enhance the quality of the generated content by evaluating the similarity between the generated results and user feedback.
% In this section, we briefly retrospect the development of DPO, which is derived from LLMs and its variants in T2I models.

\subsection{DPO in LLM Tasks}
%%%%%%%%%%%%%%%%%%%%%%%%%%%%%%%%%%%%%%%%%%%%%%%%%%%%%
Direct Preference Optimization (DPO)~\cite{rafailov2023direct} is a preference alignment method that does not require training a reward model, and is applicable for optimizing LLM. DPO optimizes the generation probabilities of paired preference data \((x_w, x_l)\) such that the preferred sample \(x_w\) has a higher probability than the non-preferred sample \(x_l\). The DPO objective can be expressed as:

% \begin{equation}
% % \begin{align}
% \begin{aligned}
% &L_{\text{DPO}} = - \mathbb{E}_{(c, x_w, x_l) \sim D} [ \log \sigma ( \beta \log \frac{p_{\theta}(x_w | c)}{p_{\text{ref}}(x_w | c)} \\
% &- \beta \log \frac{p_{\theta}(x_l | c)}{p_{\text{ref}}(x_l | c)} ) ]
% % \end{align}
% \end{aligned}
% \end{equation}
% \vspace{-3mm}
\begin{equation}
\begin{aligned}
 L =& - \mathbb{E}_{(c, x_w, x_l) \sim D} \Bigg[ \log \sigma \Bigg(\beta \log \frac{p_{\theta}(x_w | c)}{p_{\text{ref}}(x_w | c)} \\
& - \beta \log \frac{p_{\theta}(x_l | c)}{p_{\text{ref}}(x_l | c)} \Bigg) \Bigg]
\end{aligned}
\end{equation}
where \(p_{\theta}(x | c)\) and \(p_{\text{ref}}(x | c)\) represents the probability distribution generated by the DPO-trained LLM and the reference (pre-trained) model, respectively. The function \(\sigma(x)\) is the sigmoid function, and \(\beta\) controls the regularization strength. 
DPO enhances the content generated by \(p_{\theta}\) to align more closely with human preferences, thus avoiding biases and high computational costs associated with reward models in RLHF training.
% \begin{equation}
% L_{\text{DPO}}^{\text{LLM}}(\theta) = \mathbb{E}_{(y^{w}, y^{l}) \sim D} \left[ \log \sigma \left( f(y^{w}) - f(y^{l}) \right) \right]
% \label{eq:llm_dpo}
% \end{equation}

% \subsection{DPO in T2I Tasks}
% \begin{align}
% L_{\text{DPO}}^{\text{T2I}}(\theta) = & \mathbb{E}_{(x^{w}, x^{l}) \sim D, t \sim U(0, T)} \Bigg[ \log \sigma \Bigg( -\beta T \omega(\lambda_{t}) \\
% & \times \left( \| \epsilon^{w} - \epsilon_{\theta}(x^{w}, t) \|^2_{2} - \| \epsilon^{w} - \epsilon_{\text{ref}}(x^{w}, t) \|^2_{2} \right) \Bigg) \Bigg] \notag
% \end{align}

\subsection{Diffusion-DPO in T2I Tasks}
In Text-to-Image (T2I) tasks, the sampling process of diffusion models is executed step-by-step, where the objective is not to directly optimize the final generated image but to influence the denoising process at each time step \( t \), enhancing the likelihood of recovering preferred samples. Thus, the DPO objective can be extended to the diffusion process, optimizing the denoising probability distribution at each time step \( t \):
\begin{equation}
\begin{aligned}
&L_ = - \mathbb{E}_{(c, x_w, x_l) \sim D, t \sim U(0,T)} \log \sigma\bigg[ \beta \mathbb{E}_{x_{1:T}^w \sim p_{\theta}(x_{1:T}^w | x_0^w)} \\
&\log \frac{p_{\theta}(x_0^w | x_{1:T}^w)}{p_{\text{ref}}(x_0^w | x_{1:T}^w)} - \beta \mathbb{E}_{x_{1:T}^l \sim p_{\theta}(x_{1:T}^l | x_0^l)} \log \frac{p_{\theta}(x_0^l | x_{1:T}^l)}{p_{\text{ref}}(x_0^l | x_{1:T}^l)} \bigg]
\end{aligned}
\end{equation}
Here, condition text is compactness. \( x_{0:T} \) denotes the complete diffusion path, \( p_{\theta}(x_{1:T} | x_0) \) represents the diffusion process given the initial state \( x_0 \), \( p_{\theta}(x_0 | x_{1:T}) \) is the denoising probability distribution after the given diffusion trajectory \( x_{1:T} \), and \( p_{\text{ref}}(x_0 | x_{1:T}) \) refers to the corresponding distribution of the reference model. By optimizing the log probability ratio throughout the diffusion process, DPO-T2I encourages the model to generate images that align more closely with human preferences, thereby enhancing the quality of alignment in T2I tasks.

\section{Methodology}
\begin{figure*}[t]  %
	\centering
	\includegraphics[width=0.97\linewidth]{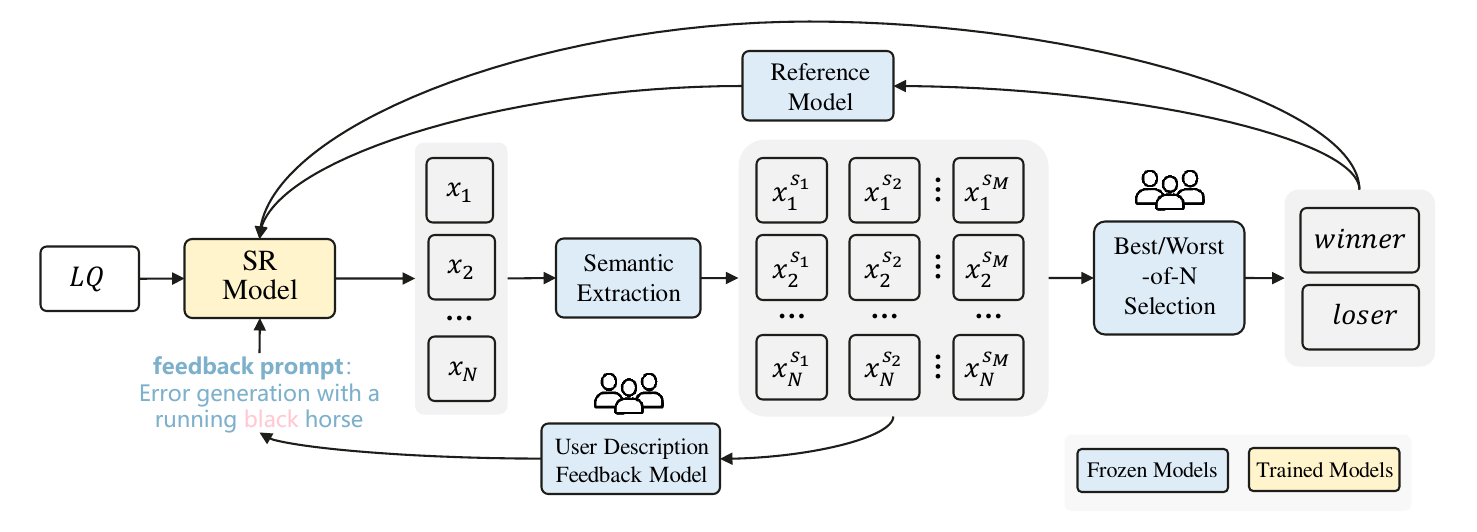}
	\caption[DSPO Framework]{The overview of proposed direct semantic preference optimization (DSPO) method.  
 % DSPO achieves finer-grained alignment of human preferences than image-level DPO.
 }
\label{fig:framework}\vspace{-1em}
\end{figure*}

\subsection{Overview}
Directly applying Diffusion-DPO to Real-ISR models degrades performance due to the dilemma between image-level preferences in DPO and pixel-level reconstruction objectives in Real-ISR. Such conflicts may make models overly sensitive to local anomalies, causing training instability and degrading generation quality.
% DSPO first samples the input LQ multiple times through a pre-trained SR model, and uses a semantic extraction model (e.g., SAM~\cite{kirillov2023segment}) to extract instance-level semantics. 
%We introduce DPO to enhance human perceptual quality by using a pairwise preference dataset, enabling the model to generate high-resolution results that align more closely with human subjective perception. We propose DSPO to address the issue where existing DPO methods in SR tasks only achieve image-level preference alignment, leading to artifacts and hallucinations in local regions, especially in high frequency texture and complex areas, which causes training instability and degrades generation quality. 
To address this issue, we propose Direct Semantic Preference Optimization (DSPO), which enhances instance-level human preference alignment by integrating semantic guidance. The overview of the proposed DSPO method is illustrated in Fig.\ref{fig:framework}. 
Specifically, to achieve finer-grained alignment, we design the semantic instance alignment strategy (detailed introduced in Sec.~\ref{sec:m1}): DSPO first generates a series of SR output with the same input LQ image using a pre-trained SR model under various settings and uses a semantic extraction model (\textit{e.g.}, SAM~\cite{kirillov2023segment}) to extract instance-level semantics. By selecting the Best/Worst-of-N, it obtains the fine-grained winners and losers, which will align with SR models through DPO. 
Additionally, to further mitigate the phenomenon of hallucination in local regions, we propose a user description feedback strategy (detailed introduced in Sec.~\ref{sec:m2}), which uses a VLM model to describe multiple instance-level samples and feeds semantically misaligned examples back to the prompt branch of the SR model by prompt injection.

\subsection{Semantic Instance Alignment Strategy}\label{sec:m1}
We propose the semantic instance alignment strategy, which aligns human preferences at the instance level to achieve finer-grained alignment.
As exhibited in Fig.~\ref{fig:framework}, based on the pre-trained SR, we first generate a series of SR output with the same input LQ image using pre-trained SR model under various settings. For each SR output, we implement an semantic extraction model (\textit{i.e.} Segment Anything (SAM)~\cite{kirillov2023segment}) to generate instances. Subsequently, we select instance-level preferred and dispreferred examples based on human evaluations and apply DPO at the instance level. The detailed implementation is as follows.

\subsubsection{Optimization Objective}
% \subsubsection{DSPO Objective}
%我们设 \( x_{\text{LQ}} \) 为输入的低分辨率图像，\( x_w \) 和 \( x_l \) 分别为用户更偏好的超分辨率（SR）图像和较劣的 SR 图像。我们通过分割模型生成 \( M \) 个语义实例区域，每个区域 \( s_m \) 表示第 \( m \) 个语义实例的掩码：
% Let \( x_{\text{LQ}} \) denote the input low-resolution image, while \( x_w \) and \( x_l \) represent the preferred SR image and dispreferred SR image, respectively. 
For an input low-resolution image $x_{\text{LQ}}$, we first generate a series of SR outputs $\{x_1, x_2, \ldots, x_N\}$. For each SR image $x_i$ generated from the pre-trained model, we generate \( M \) semantic instance regions using SAM~\cite{kirillov2023segment}: $S = \{ s_m \}_{m=1}^{M}, \quad \sum_{m=1}^{M} s_m = \mathbf{1}$,
% \begin{equation}
% S = \{ s_m \}_{m=1}^{M}, \quad \sum_{m=1}^{M} s_m = \mathbf{1},
% \end{equation}
% 其中，\( s_m \in \{0,1\}^{H \times W} \) 是二值掩码矩阵，表示语义实例的空间范围，确保所有语义实例区域构成完整的语义划分（Semantic Partitioning）。
where each region \( s_m \) represents to the mask of the \( m \)-th semantic instance, and \( s_m \in \{0,1\}^{H \times W} \) is a binary mask matrix representing the spatial coverage of the semantic instance.
% , ensuring that all semantic instance regions together form a complete segmentation.
% 在 DSPO 训练中，不同语义实例对感知质量的影响不同，因此需要对其优化贡献进行加权。我们引入 \textbf{语义重要性权重} \( w_m \)，用于平衡各语义实例的优化强度,其定义如下：
In the construction process of the optimization objective, different semantic instances have different contributions to optimization due to their varying size of areas.
% it is necessary to weight their contributions to optimization.
Therefore, we assign the weight $w_m$ for optimization intensity of each semantic instance, defined as follows:
%%%%%%%%%%%%具体而言，\( w_m \) 由语义实例 \( s_m \) 内的有效像素数归一化计算，以确保优化目标在所有语义实例间均衡分配，同时保证关键区域得到优先优化。其定义如下：
\begin{equation}
w_m = \frac{|s_m|}{\sum_{m=1}^{M} |s_m|},
\end{equation}
% 其中，\( |s_m| \) 表示语义实例 \( s_m \) 的像素数，占比越高的区域在优化过程中贡献越大，确保优化权重与语义分布相匹配。
where the term \( |s_m| \) denotes the number of pixels in the semantic instance \( s_m \). 
% Regions with the larger area contribute more significantly during optimization.

After segment, each SR image $x_i$ can be divided into different instances $\{x_i^{s_{1}}, x_i^{s_{2}}, \ldots, x_i^{s_{M}}\}$
This allows us to obtain the different instance-level image within the same instance $s_{m}$ region across different images: $\{x_1^{s_{m}}, x_2^{s_{m}}, \ldots, x_N^{s_{m}}\}$. We then perform a human preference-based Best/Worst-of-N selection and obtain $x_w^{s_{m}}$ and $x_l^{s_{m}}$, respectively. The whole image \( x_w \) of the best instance $x_w^{s_{m}}$ is defined as the preferred example, while the whole image \( x_l\) of $x_l^{s_{m}}$ is dispreferred example.
%, ensuring that the optimization weights align with the semantic distribution. 
The final optimization objective is as follows:
% \begin{equation}
% \begin{aligned}
% &\max_{\theta} \mathbb{E}_{(x_{\text{LQ}}, x_w, x_l) \sim D} \sum_{m=1}^{M} w_m \cdot [\log \sigma (\beta  \log \frac{p_\theta(x_w | x_{\text{LQ}}, s_m)}{p_{\text{ref}}(x_w | x_{\text{LQ}}, s_m)} \\
% &- \beta \log \frac{p_\theta(x_l | x_{\text{LQ}}, s_m)}{p_{\text{ref}}(x_l | x_{\text{LQ}}, s_m)} ) ]
% \end{aligned}
% \end{equation}
\begin{equation}
\begin{aligned}
&\max_{\theta} \mathbb{E}_{(x_{\text{LQ}}, x_w, x_l) \sim D} \sum_{m=1}^{M} w_m \cdot \Bigg[ \log \sigma \Bigg(\\
& \beta  \log \frac{p_\theta(x_w | x_{\text{LQ}}, s_m)}{p_{\text{ref}}(x_w | x_{\text{LQ}}, s_m)}\quad - \beta \log \frac{p_\theta(x_l | x_{\text{LQ}}, s_m)}{p_{\text{ref}}(x_l | x_{\text{LQ}}, s_m)} \Bigg) \Bigg]
\end{aligned}
\end{equation}
% \item $p_{\theta}(x | x_{\text{LQ}}, s_m)$：目标模型在语义实例 $s_m$ 内生成 x_{\text{LQ}} 图像的概率。
% \item $p_{\text{ref}}(x | x_{\text{LQ}}, s_m)$：参考模型（未经 diffusion-DPO 训练的扩散模型）在相同语义实例内的生成概率
where \( p_{\theta}(x | x_{\text{LQ}}, s_m) \) represents the probability of the target model generating an SR image $x$ within the semantic instance \( s_m \), while \( p_{\text{ref}}(x | x_{\text{LQ}}, s_m) \) represents the same for the reference (pre-trained) model.

\subsubsection{Loss Function}
%引入分割掩码 \( m_i \) 使得损失函数能够精确计算每个部分的噪声差异。在损失计算中，分割掩码的使用确保了模型能够针对每个语义信息的特性进行加权处理。这一做法使得模型在评估噪声时，能够更有效地聚焦于关键的语义信息，减少对不重要部分的干扰，从而实现更为精准的去噪效果。
% The introduction of segmentation masks \( m_i \) allows the loss function to accurately compute the noise differences for each segment. The use of these masks ensures that the model can apply weighted processing tailored to the characteristics of each semantic information segment. This approach enables the model to effectively focus on critical semantic features during noise evaluation, thereby minimizing interference from less important aspects and achieving more precise denoising results.
% For each region i, we define the new region diffusion-DPO loss as follows:
%%%%%%%%%%%%%%%%%%%%%%%%%%%%%%%%%%%%%%%%%%%%%%%%
%在扩散模型的框架下，我们进一步将目标函数转换为基于噪声预测误差的损失函数，确保优化在扩散过程的每个时间步t上逐步进行。最终的 DSPO 损失函数如下：
Referring to Diffusion-DPO~\cite{wallace2024diffusion}, the optimization objective of DPO is further reformulated into a loss function grounded in noise prediction error within the framework of diffusion models, thereby ensuring that optimization is calculated progressively at each time step \( t \) throughout the diffusion process. Therefore, our loss function can be expressed as follows:
% \begin{equation}
% \begin{aligned}
%    &L_{\text{SR}} = -\mathbb{E}_{(x_{\text{LQ}}, x_w, x_l) \sim D, t \sim U(0, T)} [ \sum_{m=1}^{M} w_m \cdot \log \sigma ( \beta T \gamma(\lambda_t) \\
%    &( (L_{\text{diff}}^\theta(x_w, t, s_m | x_{\text{LQ}}) - L_{\text{diff}}^\theta(x_l, t, s_m | x_{\text{LQ}})) - \\ 
%    &(L_{\text{diff}}^{\text{ref}}(x_w, t, s_m | x_{\text{LQ}}) + L_{\text{diff}}^{\text{ref}}(x_l, t, s_m | x_{\text{LQ}}) ))) ]
% \end{aligned}
% \end{equation}
%%%%%%%%%%%%%%%%%old_correct_version
% \begin{equation}
% \begin{aligned}
%    &L_{\text{SR}} = -\mathbb{E}_{(x_{\text{LQ}}, x_w, x_l) \sim D, t \sim U(0, T)} [ \sum_{m=1}^{M} w_m \log \sigma (- \beta T \gamma(\lambda_t) \\
%    &(L_{\text{diff}}^\theta(x_w, t, s_m | x_{\text{LQ}}) -   
%    L_{\text{diff}}^{\text{ref}}(x_w, t, s_m | x_{\text{LQ}})- \\       &(L_{\text{diff}}^\theta(x_l, t, s_m | x_{\text{LQ}}) -
%    L_{\text{diff}}^{\text{ref}}(x_l, t, s_m | x_{\text{LQ}}) 
%    ))) ] , 
% \end{aligned}
% \end{equation}
\vspace{-5mm}
\begin{equation}
\begin{aligned}
   &L_{\text{SR}} = -\mathbb{E}_{(x_{\text{LQ}}, x_w, x_l) \sim D, t \sim U(0, T)} \Bigg[ \sum_{m=1}^{M} w_m \log \sigma \Bigg( - \beta T  \\
   &\quad\times\Bigg( L_{\text{diff}}^\theta(x_w, t, s_m | x_{\text{LQ}}) - L_{\text{diff}}^{\text{ref}}(x_w, t, s_m | x_{\text{LQ}}) - \\
   &\quad \Bigg( L_{\text{diff}}^\theta(x_l, t, s_m | x_{\text{LQ}}) - L_{\text{diff}}^{\text{ref}}(x_l, t, s_m | x_{\text{LQ}}) \Bigg) \Bigg) \Bigg] ,
\end{aligned}
\end{equation}
%语义实例 $s_m$ 内的去噪误差，衡量当前模型 $\theta$ 在语义实例 $s_m$ 内的预测误差
where \( L_{\text{diff}}(x, t, s_m | x_{\text{LQ}}) \) represents the denoising error within the semantic instance \( s_m \), serving as a stability constraint during training to reduce distribution shift.
$L_{\text{diff}}^\theta$ and $L_{\text{diff}}^{\text{ref}}$ calculate the prediction error of the current model \( \theta \) and reference model, respectively, and are defined as follows:
\begin{equation}
L_{\text{diff}}^{\text{ref}}(x, t, s_m | x_{\text{LQ}}) = \gamma(\lambda_t)|| s_m \cdot (\epsilon - \epsilon_{\text{ref}}(x, t | x_{\text{LQ}})) ||^2
\end{equation}
\begin{equation}
L_{\text{diff}}^\theta(x, t, s_m | x_{\text{LQ}}) = \gamma(\lambda_t)|| s_m \cdot (\epsilon - \epsilon_\theta(x, t | x_{\text{LQ}})) ||^2
\end{equation}
where \( \epsilon_\theta(x, t | x_{\text{LQ}}) \) denotes the noise predicted by the diffusion model at time step \( t \), where \( \epsilon \) represents the true noise.  $\lambda_t$ represents the signal-to-noise ratio~\cite{kingma2021variational}, and $\gamma(\lambda_t)$ is a predefined weighting function, often set as constant~\cite{ho2020denoising,song2019generative}.
%$L_{\text{diff}}^{\text{ref}}(x, t, s_m | x_{\text{LQ}})$：参考模型的去噪误差，作为额外的稳定性约束，确保训练过程中模型不会发生分布漂移。
%\item $\epsilon_\theta(x_t, t | x_{\text{LQ}})$：扩散模型在时间步 $t$ 上预测的噪声，$\epsilon$ 代表真实噪声
%该损失函数通过语义实例级别的优化，使扩散超分辨率模型在关键区域的恢复更符合人类感知，同时借助参考模型p_ref提供稳定性约束，避免分布漂移。优化过程在每个时间步均朝向人类偏好调整，确保高语义区域的恢复质量，同时提升全局感知一致性和生成可控性，从而提高模型的泛化能力与训练稳定性。
This loss function enables optimization at the semantic instance level at each time step, enabling the SR model to align with human perception in finer-grained regions. Additionally, it leverages the reference model to prevent distribution drift. 
\subsection{User Description Feedback Strategy}
\label{sec:m2}
% 在以上策略基础上，我们进一步引入\textbf{视觉语言模型（VLM）} 以减少模型生成的语义伪像。所有 SR 结果（$o_1, o_2, ..., o_n$）在分割后分别提取各个语义实例（$o_1, s_1, ..., o_n, s_M$），并输入 VLM 进行描述分析。通过人类打分，我们识别出与输入 LQ 内容\textbf{不符的语义实例}，并收集其对应的文本描述，形成\textbf{负向提示（negative prompts）}，用于在扩散模型的采样过程中约束生成方向。该方法确保扩散模型在优化过程中减少伪生成现象，使 SR 任务更加符合语义一致性。
We further propose the user description feedback strategy to relieve the hallucination problem of generative SR models. The SR outputs $x_i$$\in$ (\(x_1, x_2, \ldots, x_N\)) are segmented to extract individual semantic instances,  %(\(o_1, s_1, \ldots, o_n, s_M\)), 
which are then analyzed through the VLM for descriptive analysis. By employing human evaluations, we identify semantic instances that do not align with the input low-quality (LQ) content and gather their corresponding text descriptions to form error generation prompts. These prompts are utilized to constrain the generation direction during the sampling process of the diffusion model to avoid hallucination. This approach ensures that the diffusion model can minimize hallucination generation throughout the optimization process, enhancing semantic consistency in the super-resolution task.
\subsection{DSPO Objectives}
The final expression for the loss function can be represented as follows:
\begin{equation}
\begin{aligned}
L_{\text{DSPO}}^{\text{}} &= \left.\sum_{m} L_{\text{}}(x_{w}, x_{l} | x_{\text{LQ}}, s_m,p_{\text{negative}}) \right.
\end{aligned}
\end{equation}
% \textbf{S-DPO结合视觉语言模型（VLM）的约束方法}通过对语义实例级别的优化显著提升了超分辨率模型的细粒度恢复能力，同时利用负向提示有效抑制了模型的伪生成问题。我们不仅增强了模型识别和修正不理想输出的能力，还促进了对人类偏好的更深入理解。
The DSPO strategy combines semantic instance alignment strategy and user description feedback strategy significantly enhancing the fine-grained recovery capability of SR models through optimization at the semantic instance level, while effectively mitigating the issue of hallucination generation using feedback description prompts. DSPO enhances the model's understanding of human preferences and improves its SR ability by addressing artifacts and hallucinations, especially in high-frequency texture and complex regions.

% \begin{equation}
% \begin{aligned}
% & L_{\text{DPO}}(x^{w}_{i}, x^{l}_{i}, m_{i}) =  \mathbb{E}_{t \sim U(0, T)} \left[ \sum_{i} \log \sigma \left( -\beta T \omega(\lambda_{t}) \right. \right. \\
% & \left. \left. \times \left( \| m_{i}(\epsilon^{w}_{i} - \epsilon_{\theta}(x^{w}_{i}), t) \|^2_{2} - \| m_{i}(\epsilon^{w}_{i} - \epsilon_{\text{ref}}(x^{w}_{i}, t) ) \|^2_{2} \right) \right) \right]
% \end{aligned}
% \end{equation}

% \begin{equation}
% \begin{aligned}
% L &= a + b \\
% + &\quad c +d +eeeeeeeeeeeeeeeeeeeeeeeeeeeeeeeeeee
% \end{aligned}
% \end{equation}
% 先验 simiao
% 方法 miaomiao
\section{Experiment}
\subsection{Experiment Setting}
\subsubsection{Training}%?是否可以这样写
%为了验证我们方法的鲁棒性，我们把DSPO分别应用在了one-step(OSEDiff)和muti-step（SD2）的sr框架上。首先在pre-trained阶段，我们在 LSDIR [21] 数据集上训练模型，模型的输入被随机crop为512x512，并使用 Real-ESRGAN [39] 的退化管道生成 LR- 训练对，共84991个lq-。其中one-step的训练参数设置和【1】一致，其中VAE 编码器、扩散网络和微调正则化器中的 LoRA 排序设置为 4，cfg设置为7.5。muti-step采用T2I的SD 2-base模型，整个受控 T2I 模型训练了 150K 次迭代，然后我们接着将其在SR的pair对数据上训练，cfg参数设置为5.5。
\paragraph{Pre-training}
To validate the robustness, we apply DSPO to both one-step and multi-step SR frameworks. For the first pre-training phase, we train both frameworks on the LSDIR~\cite{li2023lsdir} dataset. The model inputs are randomly cropped to $512\times 512$, and the LQ images are generated using the degradation pipeline from Real-ESRGAN~\cite{wang2021real}, resulting in a total of $84991$ LQ-HQ pairs. For one-step framework, the training parameters and framework are set to be consistent with OSEDiff~\cite{wu2025one}, where the LoRA rank is set to 4, and the classifier-free guidance (cfg) scale is set to 7.5. For the multi-step framework, the T2I SD 2-base\footnote{https://huggingface.co/stabilityai/stable-diffusion-2-base} is used, with the cfg scale set to 5.5. 

% 在DSPO阶段。首先，针对LSDIR的84991个lq，针对每个lq，我们会利用第一阶段情况下不同的网络参数生成4个不同的结果，比如one-step，在原先参数的基础上，我们分别修改lora为16，64，cfg分别修改为6，12的情况下得到的结果。比如muti-step，在原先参数的基础上，我们分别修改step为20和80，cfg参数修改为10.5和4.5。
\paragraph{Dataset Preparation}
In the preparation phase of the DSPO dataset, we generate four different SR results for each LQ image using different hyperparameters from the pre-training phase. Through a series of experiments, we select four sets of SR results with significant differences. Specifically, for the one-step framework, we modify the parameters based on the original settings, with $LoRA\ rank=16$, $LoRA\ rank=64$, $cfg=6$ and $cfg=12$. Similarly, for the multi-step framework, we independently set $step=20$, $step=80$, $cfg=4.5$ and $cfg=10.5$ during inference.
% 然后在得到四组结果之后，我们分为模拟计算和人工计算两种方式选择四组结果中的winner和loser的pair对。
After obtaining four different SR results, we use SAM~\cite{kirillov2023segment} to segment them into distinct instance regions. We then evaluate the same instance regions from the four results using two approaches: human annotator method and automatic Image Quality Assessment (IQA) method, which are introduced as follows.

\paragraph{Human Annotator Method} We invite ten professionals specialized in low-level tasks to perform instance-level rankings, selecting preferred and dispreferred instance image. In addition, for the user
description feedback strategy, we include the text results generated by BLIP~\cite{li2022blip} below each image in the interactive interface. If the generated text does not align with the instance image, the user will select it as the feedback textual prompt.

\paragraph{Automatic IQA Method} We first normalize the eight metrics individually (\textit{e.g.} PSNR~\cite{wang2004image}, SSIM~\cite{wang2004image}, LPIPS~\cite{zhang2018unreasonable}, DISTS~\cite{ding2020image}, 
NIQE~\cite{zhang2015feature},
MUSIQ~\cite{ke2021musiq}, MANIQA~\cite{yang2022maniqa}, and CLIPIQA~\cite{wang2023exploring}). These metrics are commonly used to assess the fidelity, perceptual quality, and alignment between generated images and their corresponding ground truths. PSNR, SSIM, and DISTS measure image similarity, while LPIPS focus on perceptual quality. NIQE, MUSIQ, MANIQA, and CLIPIQA are designed to evaluate the aesthetic and preference alignment with human perception. Then all metrics are adjusted to be a positive trend and summed to obtain an overall score, simulating human preference ratings. In addition, for the user description feedback strategy, we compute the similarity of the BLIP-generated text between the ground truth and the four candidate instance images. When the similarity is less than 0.1, we assume that the instance image generates hallucinated, and the corresponding text is selected as the feedback prompt. Due to the high cost of the human annotator method, we select the first 500 images of LSDIR, from which the top 5 largest segmented instance regions for each image are chosen to annotate, resulting in a total of 2500 instance regions for annotation, as each LQ has four corresponding SR results.

% 针对人工计算，我们邀请十位从事low-level的专业人员进行挑选sam之后instance-level human preferred LQ和inpreferred LQ。除此之外，我们会在每个图像下方增加blip的生成结果，如果专业人员觉得生成的text和图片内容不符合，则会选中作为negative prompt。

% 针对模拟计算，我们计算sam之后instance-level 生成instance与真实instance之间的PSNR↑ SSIM↑ LPIPS↓ DISTS↓ FID↓ NIQE↓ MUSIQ↑ MANIQA↑ CLIPIQA↑的归一化后加权和，以模拟人类偏好打分来获得综合评分，其中都归一化为正向趋势。除此之外，在discrimination阶段，我们使用gt计算得到的blip和四组得到的blip之间计算相似度，当相似度<0.1的时候，我们模拟其为伪生成。最终可以得到84991组lq-winner-loser-negative prompt的四元对。由于人工标注成本高，我们选择前500张图像，每个图像选取area top5的区域，共2500个图像进行标注。

\paragraph{Implementation Details.}
% DSPO的训练，我们使用 AdamW 优化器进行训练，并在 8 张NVIDIA A100 80GB GPU 上进行，batch-size设置为4，学习率是5e-5。在推理过程中，muti-step统一我们采用 间隔 DDPM 采样（spaced DDPM sampling） [45]，采样步数为 50，cfg为5.5。对于one-step β=8000，对于muti-step β=5000
During the training of DSPO, we use the AdamW optimizer~\cite{loshchilov2017decoupled} and train on 8 NVIDIA A100 80GB GPUs, with a batch size of 4 and a learning rate of 5e-5. During inference, for the multi-step framework, we adopt spaced DDPM sampling~\cite{nichol2021improved} with 50 steps and set the cfg to 5.5. We also set $\beta=8000$.
\subsubsection{Evaluation}
% 我们将我们的方法和第一阶段的pretraning的结果以及image-level diffusion-DPO进行比较。测试集分别在合成数据和真实数据上测试。合成数据集包含 3,000 张分辨率为 512×512 的图像，其 GT 图像从 DIV2K-val [2] 随机裁剪后再经过 Real-ESRGAN 退化。真实数据来自 RealSR [4] 和 DRealSR [43]，分别包含分辨率为 128×128 和 512×512 的 LR- 图像对。合成测试集与真实世界测试集的结合，能够在可控和不可预测的退化情况下评估模型，从而保证其鲁棒性和泛化能力。
We evaluate DSPO against the following baseline methods: the pre-trained method (one-step or multi-step SR frameworks), SFT, DDPO~\cite{ho2020denoising}, and Diffusion-DPO~\cite{wallace2024diffusion}. Specifically, the SFT baseline fine-tunes the pre-trained method based only on the subset of images labeled as `preferred'. The test set is evaluated on both real-world and synthetic data. For the real-world data, we use images from RealSR~\cite{cai2019toward} and DRealSR~\cite{wei2020component}, which include LQ-HQ image pairs with resolutions of $128\times128$ and $512\times512$, respectively. The synthetic dataset consists of 3000 DIV2K-val~\cite{agustsson2017ntire} GT images as HQ images, each with a resolution of 512×512. The LQ images are generated by degraded using Real-ESRGAN~\cite{wang2021real}. By applying both synthetic and real-world test sets, we evaluate the model’s performance across both controlled and unpredictable degradation conditions, thus ensuring its robustness.

% 为了评估DSPO的有效性，使用自动化偏好指标和用户研究。对于自动化偏好指标的结果，我们展示了PSNR↑ SSIM↑ LPIPS↓ DISTS↓ FID↓ NIQE↓ MUSIQ↑ MANIQA↑ CLIPIQA↑进行评估，这些是基于图像与文本的对齐模型，经过训练能够预测给定图像及其标题的人类偏好分数。此外，我们还进行了用户研究，将Diffusion-KTO与现有基准模型进行比较。在我们的用户研究中，我们要求评审判断在给定的提示下，我们邀请标注人员，从三个方面对生成的图像进行比较：
% Q1 General Preference（总体偏好）：在给LQ和LQ的条件下，你觉得哪个生成结果更加符合您的偏好？
% Q2 Visual Appeal（保真度）：在给LQ和LQ的条件下，你觉得哪个生成结果恢复的保真度更好？
% Q3 Prompt Alignment（生成度）：你觉得哪个生成结果更加符合生成？其中一张图像由我们的Diffusion-KTO模型生成，另一张由其他方法生成，且使用相同的提示。
\paragraph{Evaluation of Human Annotator Method} 
We calculate the user preference win rates (i.e., the frequency with which the human prefers images generated by DSPO) for the human annotator method. We ask annotators to compare images generated by DSPO and another method under the same LQ condition and select the image they prefer (\textit{i.e.}, `Which image do you prefer given the LQ?').
% We compare DSPO with existing baseline models under the human annotator method setting. We ask the reviewers to assess, given a specific LQ image, and provide a General Preference: "Which  result better aligns with your preference?"

\paragraph{Evaluation of Automatic IQA Method}We utilize a series of metrics to evaluate the results of automatic IQA method and other method, including PSNR~\cite{wang2004image}, SSIM~\cite{wang2004image}, LPIPS~\cite{zhang2018unreasonable}, DISTS~\cite{ding2020image}, 
NIQE~\cite{zhang2015feature}, MUSIQ~\cite{ke2021musiq}, MANIQA~\cite{yang2022maniqa}, and CLIPIQA~\cite{wang2023exploring}.

\begin{figure}[t]
  \centering
   \includegraphics[width=\linewidth]{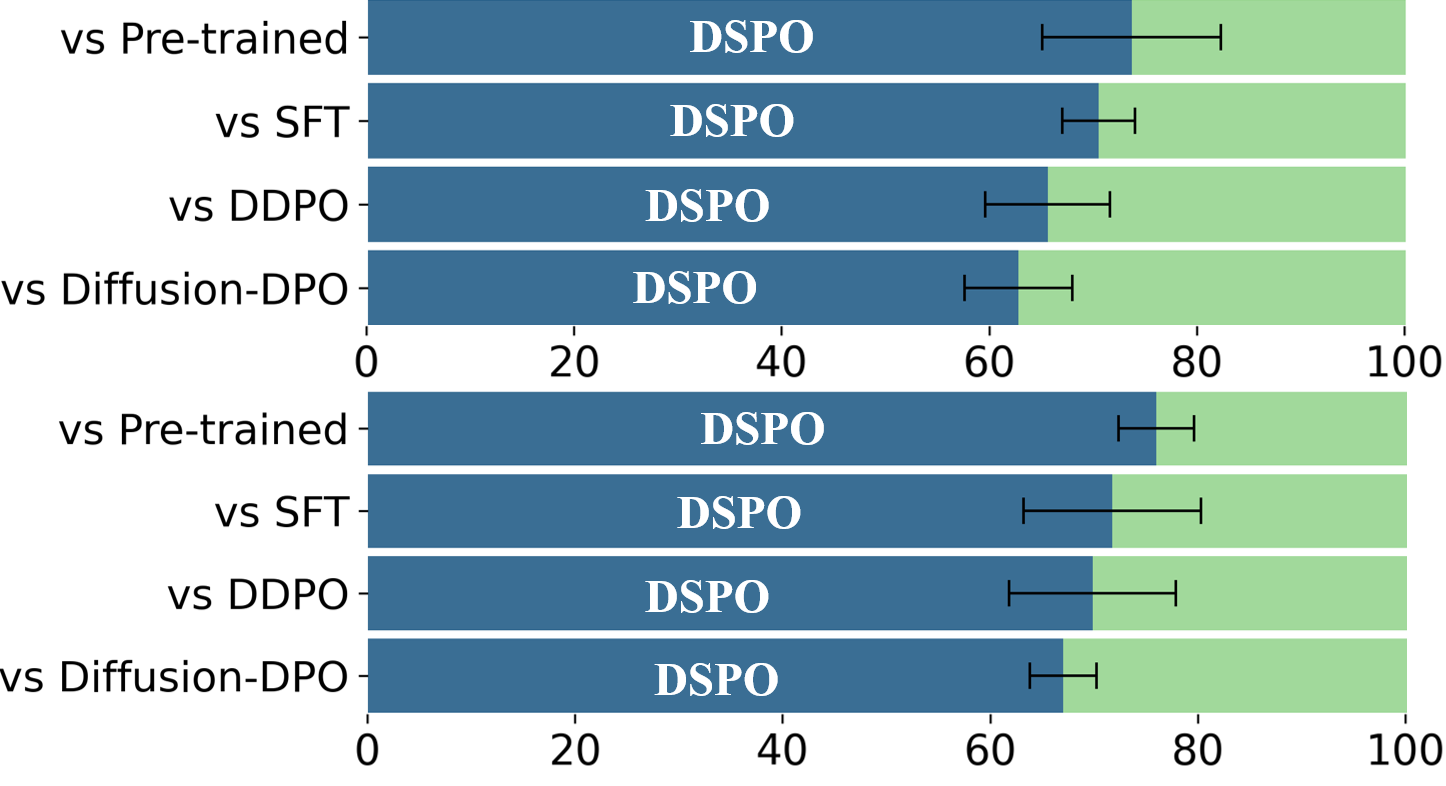}
   \vspace{-7mm}
   \caption{The user preference win rates of DSPO, compared to pre-trained method, SFT, DDPO, and Diffusion-DPO, based on the Dreal dataset (top) and the Real dataset (bottom), under human annotation. We provide the 95\% confidence interval of the win rate based on three independent annotation rounds.}
   \vspace{-2mm}
   \label{fig:human}
\end{figure}

\begin{figure*}[t]
  \centering
   \includegraphics[width=\linewidth]{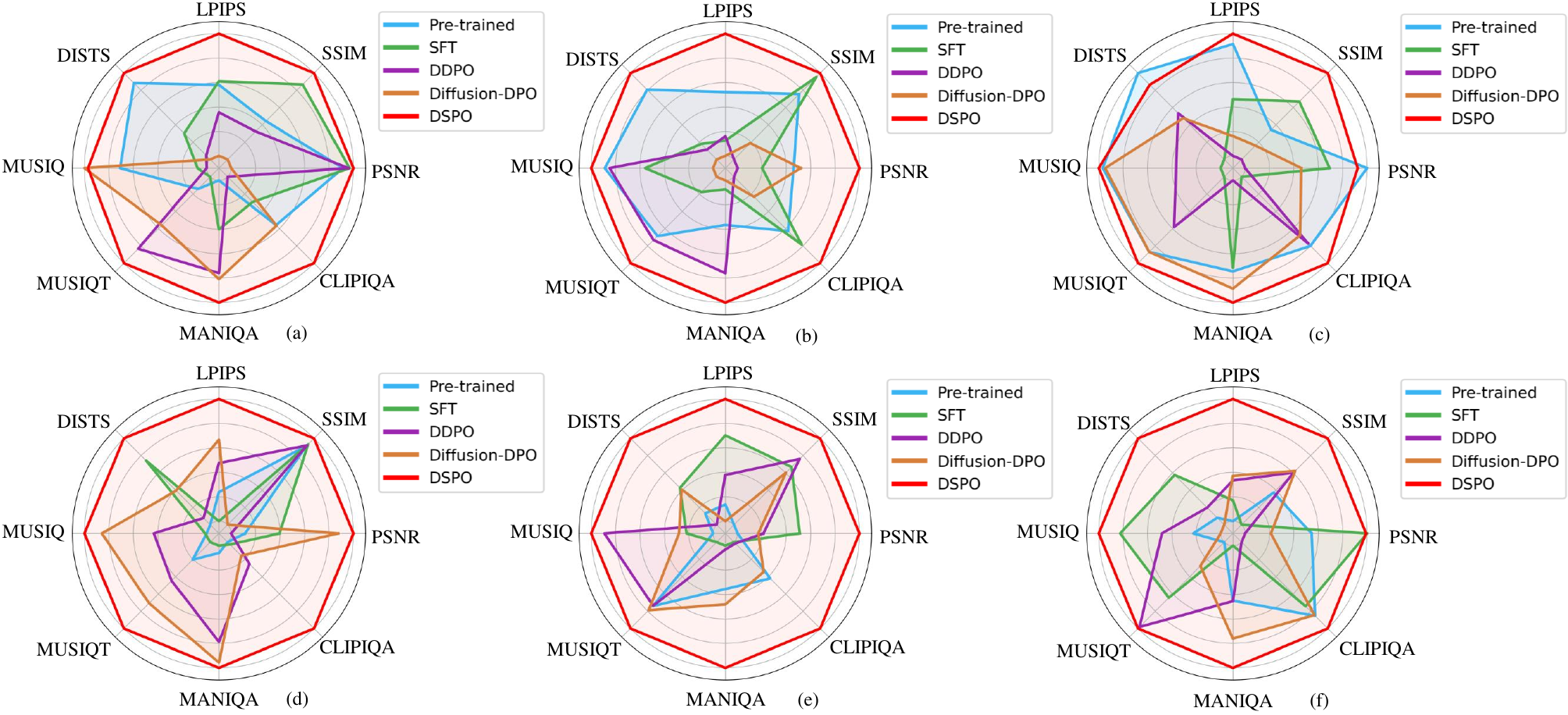}
    \vspace{-6mm}
   \caption{Quantitative comparison of DSPO with baselines on the automatic IQA Method. (a)-(c) depict the radar plots for the one-step SR framework on RealSR, DRealSR, and DIV2K-val, while (d)-(f) show radar plots for the multi-step SR framework on the same datasets. Note that all metrics are normalized and their trends are adjusted to be monotonically positive.}
   \label{fig:chart}
   \vspace{-4mm}
\end{figure*}

\begin{figure*}[t]
\centering
  % \vspace{5pt} % 增加垂直间距
  \begin{subfigure}[b]{\linewidth}
      \centering
      \includegraphics[width=0.95\linewidth]{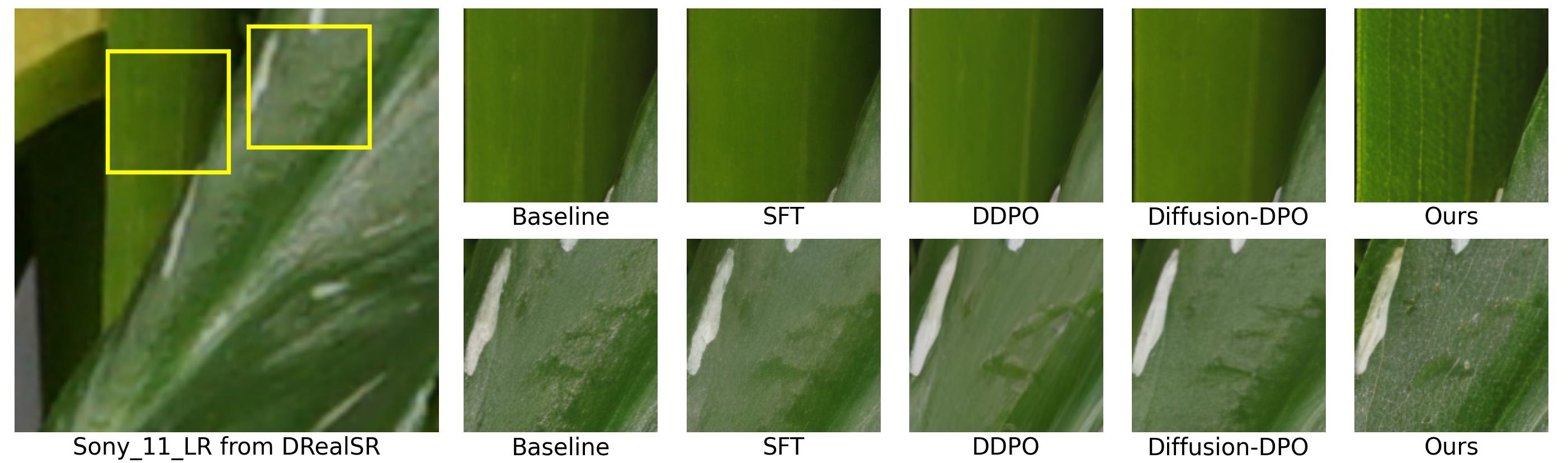}
      % \caption{Subfigure 3}
      % \label{fig:Visualization}
  \end{subfigure}
  
    \begin{subfigure}[b]{\linewidth}
      \centering
      \includegraphics[width=0.95\linewidth]{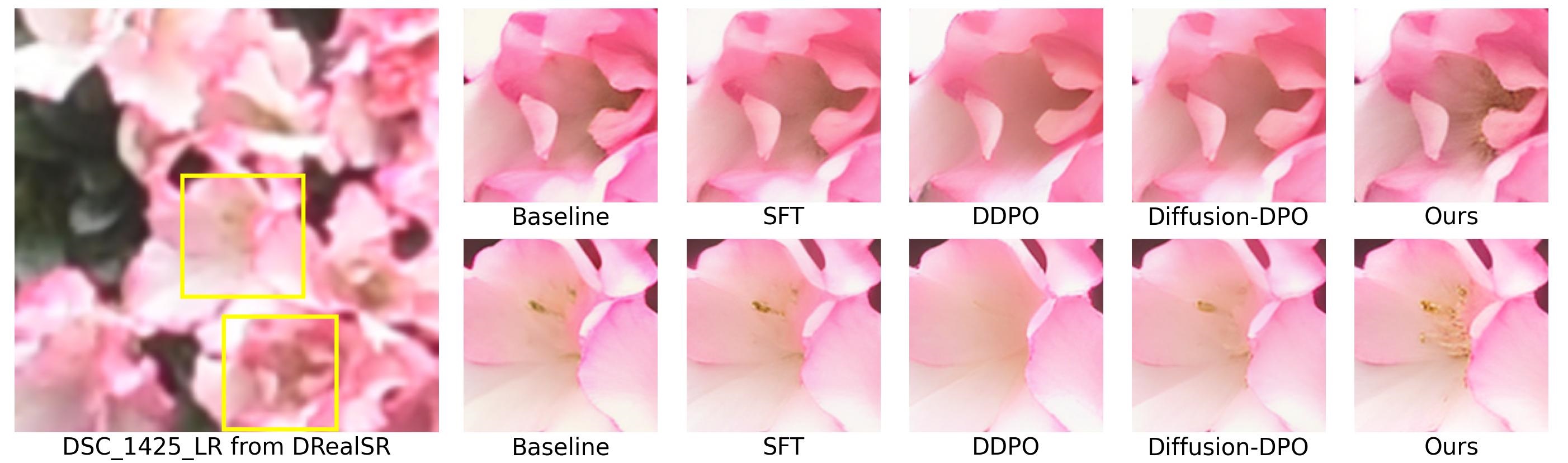}
      % \caption{Subfigure 1}
      % \label{fig:sub1}
  \end{subfigure}
  
  % \centering
  %   \begin{subfigure}[b]{\linewidth}
  %     \centering
  %     \includegraphics[width=\linewidth]{figure/human_realsr.jpg}
  %     % \caption{Subfigure 2}
  %     % \label{fig:sub2}
  % \end{subfigure}

  % \vspace{5pt} % 增加垂直间距

\vspace{-2mm}

  \caption{Visualization comparison of DSPO and other baseline methods on the human annotator method (Top), and the automatic IQA method (Bottom).}
  \label{fig:Visualization}
  \vspace{-6mm}
\end{figure*}
% \vspace{-2mm}
\subsection{Main Results}
% 图一提供了 DSPO 和pre-trained model 及相关基线模型在human 标注下的胜率based on one-step SR framework。我们邀请标注人员进行三次实验取均值，显示了error bar。可以发现，DSPO显著提升了pre-trained model的人类对其程度，最高胜率高达xx%。同时可以发现，人类评估者更偏好DSPO的方法比起SFT DDPO-DPO，这就证明DSPO的方法更能……。
\subsubsection{Preference Alignment from Annotators}
Fig.~\ref{fig:human} shows the user preference win rates of DSPO compared to the pre-trained model and other baseline models under the human annotator method, based on the one-step SR framework. Three independent annotation rounds are conducted and the 95\% confidence interval of the win rate is provided. It can be observed that DSPO significantly improves the human preference alignment of the pre-trained model, achieving an average win rate of 73.6\% and 75.6\% on the DRealSR and RealSR, respectively. Furthermore, human annotators prefer the results generated by DSPO over those produced by SFT, DDPO, and Diffusion-DPO. This indicates that for SR-based tasks, the semantic-level DSPO approach is more effective in enhancing alignment with human preferences than image-level human preference alignment methods.

% 同时，我们进行了可视化比较，如图 x 所示，更多的可视化结果放入补充材料。可以发现，我们的方法通过 instance-level 语义引导，增强对局部区域的图像超分，同时避免错误生成和伪影的出现，使整体图像更加自然、连贯。此外，DSPO 生成的图像在语义一致性上优于 baseline 方法，能够更准确地保留原始内容的结构，同时避免细节过度锐化或失真。特别是在复杂背景、人物面部结构、物体纹理以及艺术风格的保留等方面，DSPO 体现出了更强的生成能力和更高的视觉质量。
The corresponding visualization results are exhibited at the top of Fig.~\ref{fig:Visualization}, where we compare DSPO with other baseline methods. More visual results can be found in the supplementary material. It can be observed that the super-resolved images generated by DSPO align more closely with human visual perception and avoid hallucinations and artifacts, especially in image detail.
Specifically, DSPO effectively maintains the shape, edge details, and texture of target objects in a more natural manner, resulting in an overall clearer and more realistic visual appearance.

\subsubsection{Preference Alignment from Automatic IQA}
% 如图1所示，我们用雷摄图的方式提供了 DSPO 和pre-trained model 及相关基线模型在所有的自动评测指标上的结果based on one-step model还是muti-step model，所有指标进行归一化并且趋向都调整为正向。具体指标结果已放入补充材料中。可以通过观察发现，相比较于pre-trained model，我们的DSPO在所有指标上都有所提升，不管是one-step model还是muti-step model，不管是生成指标还是保真指标。具体来说，在xxx指标上提升xxx，在xxx指标上提升xxx。 而别的alignment的方法（i.e. SFT，DDDPO和Diffusion-DPO）对于基线的提升不是很明显，这是由于这些方法都是image-level，在高细节和复杂区域可能引入伪影或错误生成，导致模型对局部异常区域过于敏感，从而在训练过程中产生不确定性和波动，最终影响生成质量。而我们的方法由于进行了instance-level语义的引导，……
Fig.\ref{fig:chart} illustrates the automatic IQA model method, presented in a radar plot. It provides the results of DSPO, the pre-trained model, and related baseline models for both one-step and multi-step SR models across all evaluation metrics. Note that all metrics are normalized, and their trends are adjusted to be positive. The detailed metric results are provided in the supplementary material.
% It can be observed that, compared to the pre-trained model, DSPO achieves great improvements across all metrics, regardless of whether it is a one-step or multi-step model, and regardless of whether the metric evaluates fidelity metrics or perception metrics. 
It can be observed that, compared to the pre-trained model, DSPO achieves great improvements across all metrics, whether for one-step or multi-step models, or for fidelity or perception metrics.
Specifically, it improves by 3.34\% on the LPIPS metric on DrealSR for one-step SR and by 7.42\% on the DISTS metric on RealSR for muti-step SR.
% Furthermore, DSPO achieves more improvements compared to other alignment, demonstrating its effectiveness in enhancing image quality and fidelity.
In addition, other alignment methods (i.e., SFT, DDPO, and Diffusion-DPO) provide limited improvements over the pre-trained method due to dilemma between image-level preferences and the pixel-level reconstruction objectives of Real-ISR tasks. In contrast, DSPO leverages instance-level semantic guidance, enabling more precise capture of semantic information, reducing excessive sensitivity to local anomalies, and ultimately enhancing SR image quality.

% 同时，我们进行了可视化比较，如图 x 所示，更多的可视化结果放入补充材料。可以发现，我们的方法通过 instance-level 语义引导，能够更精准地提取和保留目标对象的关键特征，增强局部区域的细节表达，同时避免错误生成和伪影的出现，使整体图像更加自然、连贯。此外，DSPO 生成的图像在语义一致性上优于 baseline 方法，能够更准确地保留原始内容的结构，同时避免细节过度锐化或失真。特别是在复杂背景、人物面部结构、物体纹理以及艺术风格的保留等方面，DSPO 体现出了更强的生成能力和更高的视觉质量。

\begin{table*}[!t]
\centering
\resizebox{\textwidth}{!}{
\begin{tabular}{c|l|>{\centering\arraybackslash}p{1.35cm}
                >{\centering\arraybackslash}p{1.35cm}
                >{\centering\arraybackslash}p{1.35cm}
                >{\centering\arraybackslash}p{1.35cm}
                >{\centering\arraybackslash}p{1.35cm}
                >{\centering\arraybackslash}p{1.35cm}
                >{\centering\arraybackslash}p{1.35cm}
                >{\centering\arraybackslash}p{1.35cm}}
\toprule
\multicolumn{1}{l|}{\textbf{Datast}} & \textbf{Method} & \textbf{PSNR$\uparrow$} & \textbf{SSIM$\uparrow$} & \textbf{LPIPS$\downarrow$} & \textbf{DISTS$\downarrow$} & 
\textbf{NIQE$\downarrow$} & \textbf{MUSIQ$\uparrow$} & \textbf{MANIQA$\uparrow$} & \textbf{CLIPIQA$\uparrow$} \\ \midrule
\multirow{4}{*}{DrealSR}     
&Pre-trained &27.92 	&0.7835 	&0.2968 	&0.2165 	&6.4902 	&64.65 	&0.5895 	&0.6963  \\
&Perception&25.36 	&0.7157 	&0.3528 	&0.2427 	&\redbf{5.9302} 	&\redbf{68.64} 	&\redbf{0.6334} 	&\redbf{0.7080}\\
& Fidelity&\redbf{28.57} 	&\redbf{0.8024} 	&\redbf{0.2855} 	&\redbf{0.2149} 	&6.5870 	&62.76 	&0.5726 	&0.6538 \\
& United &\bluebf{27.97} 	&\bluebf{0.7919} 	&\bluebf{0.2869} 	&\bluebf{0.2150} 	&\bluebf{6.4409} 	&\bluebf{65.82} 	&\bluebf{0.6057} 	&\bluebf{0.7043} \\\midrule
\multirow{4}{*}{RealSR}
&Pre-trained&25.15 	&0.7341 	&0.2921 	&0.2128 	&5.6476 	&69.09 	&0.6331 	&0.6693 \\
&Perception&22.55 	&0.6615 	&0.3297 	&0.2270 	&\redbf{5.3484} 	&\redbf{70.19}	&\redbf{0.6633} 	&\bluebf{0.6768}  \\
& Fidelity&\redbf{25.33} 	&\redbf{0.7395} 	&\bluebf{0.2870} 	&\bluebf{0.2109} 	&5.9373 	&67.07 	&0.6177 	&0.6304\\
&United&\bluebf{25.23}	&\bluebf{0.7368} 	&\redbf{0.2826} 	&\redbf{0.2099} 	&\bluebf{5.5974} 	&\bluebf{69.42} 	&\bluebf{0.6503} 	&\redbf{0.6819}\\ \bottomrule
\end{tabular}}
\caption{The ablation study on different different optimization objectives. `All' represents the use of both perception and fidelity metrics. The best and second
best results of each metric are highlighted in \redbf{red} and \bluebf{blue}, respectively.}
\label{tab:abmetric}
\end{table*}

% \subsubsection{Rank}

\begin{table*}[!t]
\centering
\begin{tabular}{c|>{\centering\arraybackslash}p{1.4cm}
                >{\centering\arraybackslash}p{1.4cm}
                >{\centering\arraybackslash}p{1.4cm}
                >{\centering\arraybackslash}p{1.4cm}
                >{\centering\arraybackslash}p{1.4cm}
                >{\centering\arraybackslash}p{1.4cm}
                >{\centering\arraybackslash}p{1.4cm}
                >{\centering\arraybackslash}p{1.4cm}}
\toprule
\textbf{Method} & \textbf{PSNR$\uparrow$} & \textbf{SSIM$\uparrow$} & \textbf{LPIPS$\downarrow$} & \textbf{DISTS$\downarrow$} & 
\textbf{NIQE$\downarrow$} & \textbf{MUSIQ$\uparrow$} & \textbf{MANIQA$\uparrow$} & \textbf{CLIPIQA$\uparrow$} \\ \midrule
     
Pre-trained &27.92 	&0.7835 	&0.2968 	&0.2165 	&6.4902 	&64.65 	&0.5895 	&0.6963  \\ \cmidrule{1-9}
\multirow{2}{*}{M1}   &27.94 	&0.7888 	&0.2930 	&0.2144 &6.4451 	&65.10 	&0.6005 	&0.7088\\
&\textcolor{red}{+0.02} 	&\textcolor{red}{+0.0053}	&\textcolor{red}{-0.0038} 	&\textcolor{red}{-0.0021} 	&\textcolor{red}{-0.0451}	&\textcolor{red}{+0.45} 	&\textcolor{red}{+0.0110} 	&\textcolor{red}{+0.0125}\\ \cmidrule{1-9}
\multirow{2}{*}{M1+M2} &27.97 	&0.7919 	&0.2889 	&0.2150 	&6.4409 	&65.82 	&0.6057 	&0.7043\\
&\textcolor{red}{+0.05} 	&\textcolor{red}{+0.0084} 	&\textcolor{red}{-0.0079} 	&\textcolor{red}{-0.0015} 	&\textcolor{red}{-0.0493} 	&\textcolor{red}{+1.17}	&\textcolor{red}{+0.0162} 	&\textcolor{red}{+0.0080}
\\
% \midrule
% \multirow{5}{*}{RealSR}
% & One-step &25.15 	&0.7341 	&0.2921 	&0.2128 	&5.6476 	&69.09 	&0.6331 	&0.6693 \\
% & M1\\
% & M1+M2&\\
\bottomrule
\end{tabular}
% \vspace{-3mm}
\caption{The ablation study on different strategy. `M1' represents the semantic instance alignment strategy  and `M2' represents the user description feedback strategy. The red-highlighted values indicate improved performance  compared to the pre-trained method.}
\label{tab:abmoudle}
\vspace{-2mm}
\end{table*}

\begin{figure}[!t]
  \centering
   \includegraphics[width=\linewidth]{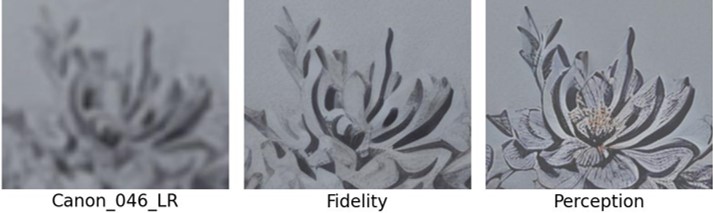}
   \vspace{-6mm}
   \caption{The visualization of  DSPO based on perception metrics and fidelity metrics, respectively.}
   \vspace{-6mm}
   \label{fig:ablation}
\end{figure}
We conduct a visual comparison, as exhibited in the middle and the bottom of Fig.~\ref{fig:Visualization}, with more results provided in the supplementary material. It can be observed that DSPO, through instance-level semantic guidance, enhances super-resolution in local regions while avoiding hallucinations and artifacts, resulting in a more natural and coherent overall LQ image. 
Moreover, the SR images generated by DSPO achieve better semantic consistency than the baseline methods, accurately preserving the structural integrity of the LQ while avoiding excessive sharpening or distortion of details. 
In particular, DSPO demonstrates superior generation capability in handling complex backgrounds, facial structures, object textures, and artistic style preservation.

\subsection{Ablation}
\paragraph{Perception and Fidelity Metric}
%为了分析不同 DSPO 优化目标对超分辨率的影响，我们分别使用 Perceptual Metric 和 Fidelity Metric，模拟人工从感知质量和保真度两个角度进行评判，并进行 DSPO 实验，结果如表 1 所示。
To analyze the impact of different DSPO optimization objectives on super-resolution, we conduct experiments using perception Metric (\textit{i.e.} NIQE, MUSIQ, MANIQA, and CLIPIQA) and fidelity metric (\textit{i.e.} PSNR, SSIM, LPIPS, and DISTS), simulating human evaluation from both perceptual quality and fidelity perspectives. The results are presented in Table~\ref{tab:abmetric}
% 实验结果表明，在 DSPO 过程中，优化 Perceptual Metric 可显著提升感知质量（Perceptual），优化 Fidelity Metric 则有效提高保真度（Fidelity），不同优化目标直接影响相应的生成效果。
Results show that DSPO optimization improves perceptual quality with the Perception metric and enhances fidelity with the fidelity metric, with each optimization objective directly influencing the corresponding generation outcomes.
% 同时，我们对两个实验的结果进行了可视化分析，如图 11 所示。可以观察到，在 DSPO 的引导下，基于 Perceptual Metric 优化的结果虽然在一定程度上牺牲了保真度，但生成的花蕊极为逼真，提升了视觉细节的丰富性。相比之下，基于 Fidelity Metric 优化的结果更加贴近 GT，能够准确还原原始内容的结构和细节。DSPO 在不同优化目标下展现出灵活的适应性，能够根据优化策略的不同，生成既符合感知偏好又具备高保真度的超分辨率图像，充分体现了其在超分任务中的有效性。
We also show visualization in Fig.~\ref{fig:ablation}. While optimizing perception metrics sacrifices some fidelity, it produces highly realistic stamens, enriching visual details. In contrast, optimizing the fidelity metric yields results closer to the GT, accurately preserving structural integrity. DSPO demonstrates adaptability across different optimization objectives, highlighting its effectiveness in super-resolution tasks.

\paragraph{Different Strategy}
% 为了分析 semantic instance alignment strategy 和 user description feedback strategy 的作用，我们进行了消融实验，如表2所示。实验结果表明，semantic instance alignment strategy 能显著提升人类偏好对齐、减少伪影并优化实例级别表现。在此基础上，加入 user description feedback strategy 后，生成性能进一步提升，优化效果更佳，两个模块的联合效果最优。
To analyze the effects of the semantic instance alignment strategy and the user description feedback strategy, we conduct an ablation study, as exhibited in Table~\ref{tab:abmoudle}. The experimental results demonstrate that the semantic instance alignment strategy significantly improves preference, and reduces artifacts and hallucination. Incorporating the user description feedback strategy further enhances the image super-resolution performance, leading to more refined results. The combination of both strategies achieves the best overall performance.

\section{Conclusion}
%We propose a novel approach to Real-ISR by integrating human preference alignment technology. 
This paper presents Direct Semantic Preference Optimization (DSPO), a novel framework that pioneers human preference alignment in Real-ISR.
%Traditional diffusion-based super-resolution models, due to the lack of human involvement, still have a gap between the optimization objectives and human perception, which can lead to artifacts, hallucinations, and potentially harmful generations.
To address the dilemma between image-level preference of DPO and pixel-wise preference alignment, our method introduces two key innovations:
%DSPO aims to align human preferences deeply by integrating semantic guidance to mitigate local artifacts and hallucinations, enhancing the perceptual quality of the generated images. 
%Specifically, DSPO utilizes a 
(1) A semantic instance alignment strategy that optimizes semantic preference learning at the instance level to achieve finer-grained alignment, and (2) a user description feedback strategy that injects user-selected semantic hallucination texts as prompts. Comprehensive experiments demonstrate DSPO's effectiveness across both one-step and multi-step frameworks, achieving significant improvements in perceptual quality and fidelity metrics. As the first principled integration of human preference learning with SR optimization, our work establishes a new paradigm for developing visually coherent and human-aligned SR systems.

% xudong
{
    \small
    \bibliographystyle{ieeenat_fullname}
    \bibliography{main}
}

\end{document}